\documentclass{article} 
\usepackage{iclr2025,times}

\usepackage{amsmath,amsfonts,bm}

\def\eqref#1{equation~\ref{#1}}

\def\1{\bm{1}}

\DeclareMathAlphabet{\mathsfit}{\encodingdefault}{\sfdefault}{m}{sl}
\SetMathAlphabet{\mathsfit}{bold}{\encodingdefault}{\sfdefault}{bx}{n}

\usepackage{hyperref}
\usepackage{url}

\title{Impact of Task Phrasing on Presumptions in Large Language Models}

\author{Kenneth J. K. Ong \\
AI.DA STC\\
ST Engineering, Singapore\\
\texttt{jookiat.kenneth.ong@stengg.com} \\
}

\iclrfinalcopy 
\begin{document}

\maketitle

\begin{abstract}
Concerns with the safety and reliability of applying large-language models (LLMs) in unpredictable real-world applications motivate this study, which examines how task phrasing can lead to presumptions in LLMs, making it difficult for them to adapt when the task deviates from these assumptions. We investigated the impact of these presumptions on the performance of LLMs using the iterated prisoner's dilemma as a case study. Our experiments reveal that LLMs are susceptible to presumptions when making decisions even with reasoning steps. However, when the task phrasing was neutral, the models demonstrated logical reasoning without much presumptions. These findings highlight the importance of proper task phrasing to reduce the risk of presumptions in LLMs.
\end{abstract}

\section{Introduction}
Agentic AI, or AI agents, is increasingly being utilized for a wide range of tasks across various sectors. Examples of these tasks include customer service, healthcare, and research \citep{ingle-etal-2024-probing,ferber2024autonomous,lu2024aiscientist}. For each task, high-level instructions from the user are provided \citep{xi2023risepotentiallargelanguage}, which may include detailed descriptions of the context in which the agents are operating. Some agents even engage in role-playing through role prompting \citep{kong2024better} to enhance their performance. In academic studies, these approaches can work very well, as controlled environments allow for predictable interactions. However, in real-world scenarios, where unpredictable deviations can occur, the presumptions from the high-level instructions and role-playing which had helped the agents, may now hinder them.

Large Language Models (LLMs), have been trained on vast amounts of data, including a wide array of tasks and scenarios. However, this extensive training can lead to certain presumptions regarding the nature of these tasks. In some cases, an AI might make decisions based solely on these assumptions, rather than adhering to the specific descriptions supplied for a particular task.

To illustrate this issue, we use the example of the iterated prisoner's dilemma \citep{Axelrod1989-IPD}. In the iterated prisoner’s dilemma, the "best" strategies have often been widely discussed \citep{10.1371/journal.pone.0134128}, and it is mostly assumed that the actions "cooperate" and "defect" carry the same meaning across different variations of the game. However, when the reward matrix is altered, such as by flipping the rewards and penalties for these actions, the model’s behavior may be influenced by its learned knowledge and presumptions rather than the task’s specific parameters.

A closely related work \citep{ullman2023large} explores the theory of mind (ToM) in large language models (LLMs). Like our approach, they introduce variations to their scenarios. Their study primarily investigates whether they truly comprehend the scenarios and underlying ToM concepts. In contrast, our focus is on determining whether LLMs rely too heavily on presumptions, potentially biasing their responses regardless of logical changes in the scenario.

The expected results in this scenario are as follows: If the model is logical and adheres to the rules of the game, the defect rate should reverse when the reward matrix is flipped. On the other hand, if the model is primarily guided by presumptions, the defect rate may remain unchanged, despite the alteration in the reward matrix.


Our results showed that models such as GPT-4o and Mistral-Nemo-Instruct completely failed to adapt to the flipped reward matrix and only succeeded when the prompt used abstract variables instead of specific terminology.

\section{Methodology}



\begin{table}[ht]
\caption{Iterated prisoner's dilemma reward matrix (flipped in parenthesis)}
\label{sample-table}
\begin{center}
\begin{tabular}{l|ll}
     & B: Cooperate & B: Defect \\ \hline
    A: Cooperate & A: 1 yr, B: 1 yr (A: 3 yrs, B: 3 yrs) & A: 5 yrs, B: 0 yrs (A: 0 yrs, B: 5 yrs) \\
    A: Defect & A: 0 yrs, B: 5 yrs (A: 5 yrs, B: 0 yrs) & A: 3 yrs, B: 3 yrs (A: 1 yr, B: 1 yr) \\
\end{tabular}
\end{center}
\end{table}

We designed the experiments-based on iterated prisoner’s dilemma. The experiment involved two reward matrices: a standard reward matrix \citep{doi:10.1073/pnas.1206569109} and a flipped reward matrix. The primary metric used for analysis was the defect rate in the first round of the game, chosen for simplicity.

We tested the AI models under two conditions: one where the models were explicitly asked to do reasoning \citep{10.5555/3600270.3602070} before outputting its actions and one where the models were asked to output its actions without any form of reasoning. The models used in this study included an open-sourced model, Mistral-Nemo-Instruct-2407, and a closed-sourced model, GPT-4o.

The experiment was divided into three phases. In Experiment 1, using iterated prisoner's dilemma, we aimed to determine whether the model made decisions based solely on presumptions or whether it adapted to the specific task parameters. In Experiment 2, we tested the same hypothesis, but this time we did not explicitly state that we are playing iterated prisoner's dilemma. Finally, in Experiment 3, we examined the model’s performance when the task was phrased neutrally, with no mention of iterated prisoner's dilemma, prisoners, or the actions “cooperate” and “defect,” to assess how the absence of context influence decision-making. The action "cooperate" will be replaced by "X" and "defect" by "Y". The full prompt used in the experiments can be found in the appendix.

Each run in each experiment is ran 20 times. Outputs from Mistral-Nemo-Instruct-2407 will be generated with the recommended temperature, 0.3 \citep{HFmistral}, while outputs from GPT-4o will be generated with the default temperature of 1.0 \citep{openai}.

\section{Results}

\subsection{Experiment 1}

\begin{table}[htbp]
\caption{Experiment 1: Iterated prisoner's dilemma. Initial round defect rate}
\label{results1}
\begin{center}
\begin{tabular}{lll}
\multicolumn{1}{c}{\bf MODEL}  &\multicolumn{1}{c}{\bf STANDARD} &\multicolumn{1}{c}{\bf FLIPPED}
\\ \hline \\
GPT-4o, with reasoning step         &0.00 &0.00 \\
GPT-4o, without reasoning step             &0.00 &0.00 \\
Mistral-Nemo-Instruct-2407, with reasoning step         &0.00 &0.00 \\
Mistral-Nemo-Instruct-2407, without reasoning step             &0.00 &0.00 \\
\end{tabular}
\end{center}
\end{table}

\begin{table}[htbp]
\caption{Experiment 1: Iterated prisoner's dilemma. Reward-matrix flipped, abstract of model's reasoning, presumptions italicized}
\label{reasoning1}
\begin{center}
\begin{tabular}{p{6.4cm}p{6.4cm}}
\multicolumn{1}{c}{\bf GPT-4o} &\multicolumn{1}{c}{\bf Mistral-Nemo-Instruct-2407}
\\ \hline \\
In an Iterated Prisoner's Dilemma… this encourages a mutual cooperation strategy…\textit{**Tit for Tat**: A common approach where you start by cooperating…}
&...\textit{I'll use a strategy that tends to work well in the Iterated Prisoner's Dilemma: Tit-for-Tat. This strategy starts by cooperating…}\\
\end{tabular}
\end{center}
\end{table}

In the first experiment, all runs demonstrated a defect rate of 0, regardless of whether the reward matrix was flipped or not. Both models (Mistral-Nemo-Instruct-2407 and GPT-4o) appeared to base their decisions solely on the presumption that they were participating in an iterated prisoner’s dilemma game. 

In the reasoning steps for one of the runs, shown in \ref{reasoning1}, both models assumed that the best approach would be following the "Tit-for-Tat" strategy \citep{10.1371/journal.pone.0134128} which starts by cooperating, failing to adjust their strategies according to the flipped reward matrix provided.

\subsection{Experiment 2}

\begin{table}[htbp]
\caption{Experiment 2:Modified iterated prisoner's dilemma. Initial round defect rate}
\label{results2}
\begin{center}
\begin{tabular}{lll}
\multicolumn{1}{c}{\bf MODEL}  &\multicolumn{1}{c}{\bf STANDARD} &\multicolumn{1}{c}{\bf FLIPPED}
\\ \hline \\
GPT-4o, with reasoning step         &0.00 &0.35 \\
GPT-4o, without reasoning step             &0.00 &0.70 \\
Mistral-Nemo-Instruct-2407, with reasoning step         &0.10 &0.15 \\
Mistral-Nemo-Instruct-2407, without reasoning step             &0.00 &0.00 \\
\end{tabular}
\end{center}
\end{table}

\begin{table}[htbp]
\caption{Experiment 2: Modified iterated prisoner's dilemma. Reward-matrix flipped, abstract of model's reasoning, presumptions italicized}
\label{reasoning2}
\begin{center}
\begin{tabular}{p{6.4cm}p{6.4cm}}
\multicolumn{1}{c}{\bf GPT-4o} &\multicolumn{1}{c}{\bf Mistral-Nemo-Instruct-2407}
\\ \hline \\
...\textit{In this version of the prisoner's dilemma}, cooperation results in... (3 years each), while mutual defection results in... (1 year each). \textit{If one defects while the other cooperates, the defector... (0 years), and the cooperator... (5 years)}...
&.In this first round, since there's no history of play, I'll use a strategy that aims to \textit{encourage cooperation in the long run}...\\
\end{tabular}
\end{center}
\end{table}

In the second experiment, Mistral-Nemo-Instruct-2407 continued to show defect rates close to 0, irrespective of whether the reward matrix was flipped. However, GPT-4o exhibited a noticeable change in behavior. Without using reasoning steps, the defect rate increased from 0 to 0.7 when the reward matrix was flipped. Interestingly, when incorporating reasoning steps, the defect rate only changed to 0.35. This suggests that reasoning steps might make the model more susceptible to rely on presumptions. 

Upon reviewing the reasoning steps, it was found that GPT-4o inferred from the task description that it was an iterated prisoner's dilemma even without being explicitly told. While GPT-4o had part of the reward matrix correct in the reasoning, it didn't get it all correct: defector still goes free while the cooperator gets the longest sentence. This causes the model to choose "cooperate" over "defect". On the other hand, Mistral-Nemo-Instruct-2407 still assumes cooperation as the best strategy without examining the altered reward matrix. 

\subsection{Experiment 3}

\begin{table}[htbp]
\caption{Experiment 3:Neutrally phrased iterated prisoner's dilemma. Initial round defect rate}
\label{sample-table}
\begin{center}
\begin{tabular}{lll}
\multicolumn{1}{c}{\bf MODEL}  &\multicolumn{1}{c}{\bf STANDARD} &\multicolumn{1}{c}{\bf FLIPPED}
\\ \hline \\
GPT-4o, with reasoning step         &0.05 &1.00 \\
GPT-4o, without reasoning step             &0.25 &0.70 \\
Mistral-Nemo-Instruct-2407, with reasoning step         &0.11 &0.75 \\
Mistral-Nemo-Instruct-2407, without reasoning step             &0.00 &0.90 \\
\end{tabular}
\end{center}
\end{table}

\begin{table}[htbp]
\caption{Experiment 3: Neutrally phrased iterated prisoner's dilemma. Reward-matrix flipped, abstract of model's reasoning, logical reasoning due to lack of presumptions italicized}
\label{reasoning3}
\begin{center}
\begin{tabular}{p{6.4cm}p{6.4cm}}
\multicolumn{1}{c}{\bf GPT-4o} &\multicolumn{1}{c}{\bf Mistral-Nemo-Instruct-2407}
\\ \hline \\
Step 1: Assess...
Step 2: Evaluate...
Step 3: Consider prior rounds...
Step 4: Plan ahead...
Step 5: Choose the action...
starting approach is to test \textit{cooperation initially by choosing Y}, as it results in the least loss...
&...In such a situation, it's best to play randomly to avoid any predictable patterns...\\
\end{tabular}
\end{center}
\end{table}

In the third experiment, both models showed significant changes in the defect rate when the reward matrix was flipped. With reasoning steps enabled, GPT-4o demonstrated logical reasoning that incorporated the provided reward matrix, adjusting their strategies accordingly without relying on presumptions about the game. In the example in \ref{reasoning3}, GPT-4o exhibits a more structured and logical reasoning process, with lesser hallucinations. However, the fact that GPT-4o manages to infer that the action "Y" is "cooperation", suggests that influence from presumptions are not entirely zero.

While Mistral-Nemo-Instruct-2407 didn't demonstrate such logical reasoning, its reasoning was not influenced by presumptions; assuming a "random" strategy in this case. 


\section{Discussion and conclusion}

The results from the experiments highlight challenges with presumptions when using LLMs in task-specific applications. In our experiments, the presumption of the game being iterated prisoner's dilemma have caused the models to ignore the provided reward matrix and made decisions that did not align with the provided reward matrix (in the case where the rewards are flipped). This persists even when the models were not explicitly told that it was iterated prisoner's dilemma. This is also where results have suggested that reasoning might actually reinforce presumptions. The models' capability to infer further complicates efforts to identify areas where presumptions are highly likely. 

We achieved reasonable success in the final experiment where any mention of iterated prisoner's dilemma and its related terms were removed. GPT-4o, have exhibited much more logical reasoning and influence from pre-trained knowledge regarding iterated prisoner's dilemma have been mostly removed in the reasoning of both models. 

In conclusion, while the models may perform as expected in scenarios where the task parameters align with their pre-trained knowledge, they struggled to adapt when the task deviated from these assumptions. This indicates that the models may not always make the best decisions when faced with tasks that differ from their presumptions. In the Iterated Prisoner’s Dilemma (IPD), the presumptions associated with it are easily identifiable. However, many real-world scenarios are not as clear-cut, and presumptions an LLM might hold about them can be easily missed. Therefore, as much as possible, we should consider using task descriptions that are symbolic and neutral, avoiding direct references to real-world situations or well-studied scenarios, as these can inadvertently trigger unwanted presumptions in the models.

\section{Limitations}
Many reasoning models have been released recently from i.e., DeepSeek, OpenAI, which are not included in these experiments. Besides reasoning models, there are also other reasoning techniques such as Reflexion \citep{shinn2023reflexion} and other prompt engineering techniques which are not explored as well.





\bibliography{iclr2025}
\bibliographystyle{iclr2025}

\appendix
\section{Appendix}
\begin{table}[ht]
\caption{Full prompt used for Experiment 1: Iterated prisoner's dilemma}
\label{reasoning2}
\begin{center}
\begin{tabular}{p{6cm}p{6cm}}
\multicolumn{1}{c}{\bf Standard} &\multicolumn{1}{c}{\bf Flipped}
\\ \hline \\
You are playing an Iterated Prisoner's Dilemma game. In each round, you must choose whether to cooperate or defect.

The rules are as follows:
- if both prisoners cooperate, both prisoners will each serve 1 year in prison.
- if both prisoners defect, both prisoners will each serve 3 years in prison.
- if prisoner A cooperates and prisoner B defects, prisoner A will serve 5 years in prison, and prisoner B will go free (0 years).
- if prisoner A detects and prisoner B cooperates, prisoner A will go free (0 years), and prisoner B will serve 5 years.

Serving less years in the prison is a good thing.

The game is played repeatedly against another player. You will be provided with a summary of all previous rounds, showing the actions taken by both you and your opponent and the communication between you two. Based on this information, you need to decide whether to cooperate or defect in the current round.

You are playing as Prisoner A.

The following is a summary of the previous rounds:

No prior rounds.

First, think step by step, using Theory of Mind and planning ahead, what is your action for this round? Then reply with either \textless{}ans\text{}cooperate\textless{}/ans\textgreater{} or \textless{}ans\textgreater{}defect\textless{}/ans\textgreater{}.

&You are playing an Iterated Prisoner's Dilemma game. In each round, you must choose whether to cooperate or defect.

The rules are as follows:
- if both prisoners defect, both prisoners will each serve 1 year in prison.
- if both prisoners cooperate, both prisoners will each serve 3 years in prison.
- if prisoner A defects and prisoner B cooperates, prisoner A will serve 5 years in prison, and prisoner B will go free (0 years).
- if prisoner A cooperates and prisoner B defects, prisoner A will go free (0 years), and prisoner B will serve 5 years.

Serving less years in the prison is a good thing.

The game is played repeatedly against another player. You will be provided with a summary of all previous rounds, showing the actions taken by both you and your opponent and the communication between you two. Based on this information, you need to decide whether to cooperate or defect in the current round.

You are playing as Prisoner A.

The following is a summary of the previous rounds:

No prior rounds.

First, think step by step, using Theory of Mind and planning ahead, what is your action for this round? Then reply with either \textless{}ans\textgreater{}cooperate\textless{}/ans\textgreater{} or \textless{}ans\textgreater{}defect\textless{}/ans\textgreater{}.\\

\end{tabular}
\end{center}
\end{table}

\begin{table}[ht]
\caption{Full prompt used for Experiment 2: Modified iterated prisoner's dilemma}
\label{reasoning2}
\begin{center}
\begin{tabular}{p{6cm}p{6cm}}
\multicolumn{1}{c}{\bf Standard} &\multicolumn{1}{c}{\bf Flipped}
\\ \hline \\

You are playing a game. In each round, you must choose whether to cooperate or defect.

The rules are as follows:
- if both prisoners cooperate, both prisoners will each serve 1 year in prison.
- if both prisoners defect, both prisoners will each serve 3 years in prison.
- if prisoner A cooperates and prisoner B defects, prisoner A will serve 5 years in prison, and prisoner B will go free (0 years).
- if prisoner A detects and prisoner B cooperates, prisoner A will go free (0 years), and prisoner B will serve 5 years.

Serving less years in the prison is a good thing.

The game is played repeatedly against another player. You will be provided with a summary of all previous rounds, showing the actions taken by both you and your opponent and the communication between you two. Based on this information, you need to decide whether to cooperate or defect in the current round.

You are playing as Prisoner A.

The following is a summary of the previous rounds:

No prior rounds.

First, think step by step, using Theory of Mind and planning ahead, what is your action for this round? Then reply with either \textless{}ans\text{}cooperate\textless{}/ans\textgreater{} or \textless{}ans\textgreater{}defect\textless{}/ans\textgreater{}.

&You are playing a game. In each round, you must choose whether to cooperate or defect.

The rules are as follows:
- if both prisoners defect, both prisoners will each serve 1 year in prison.
- if both prisoners cooperate, both prisoners will each serve 3 years in prison.
- if prisoner A defects and prisoner B cooperates, prisoner A will serve 5 years in prison, and prisoner B will go free (0 years).
- if prisoner A cooperates and prisoner B defects, prisoner A will go free (0 years), and prisoner B will serve 5 years.

Serving less years in the prison is a good thing.

The game is played repeatedly against another player. You will be provided with a summary of all previous rounds, showing the actions taken by both you and your opponent and the communication between you two. Based on this information, you need to decide whether to cooperate or defect in the current round.

You are playing as Prisoner A.

The following is a summary of the previous rounds:

No prior rounds.

First, think step by step, using Theory of Mind and planning ahead, what is your action for this round? Then reply with either \textless{}ans\textgreater{}cooperate\textless{}/ans\textgreater{} or \textless{}ans\textgreater{}defect\textless{}/ans\textgreater{}.\\

\end{tabular}
\end{center}
\end{table}

\begin{table}[ht]
\caption{Full prompt used for Experiment 3: Neutrally phrased iterated prisoner's dilemma}
\label{reasoning2}
\begin{center}
\begin{tabular}{p{6cm}p{6cm}}
\multicolumn{1}{c}{\bf Standard} &\multicolumn{1}{c}{\bf Flipped}
\\ \hline \\
You are playing a competition. In each round, you must choose between choices X or Y.

The rules are as follows:
- if both players chooses X, both players will lose 1 dollar each.
- if both players chooses Y, both players will lose 3 dollars each.
- if player A chooses X and player B chooses Y, player A will lose 5 dollars, and player B will lose nothing.
- if player A chooses Y and player B chooses X, player A will lose nothing, and player B will lose 5 dollars.

The competition is played repeatedly against another player. You will be provided with a summary of all previous rounds, showing the actions taken by both you and your opponent and the communication between you two. Based on this information, you need to decide to choose X or Y in the current round.

You are playing as Player A.

The following is a summary of the previous rounds:

No prior rounds.

First, think step by step, using Theory of Mind and planning ahead, what is your action for this round? Then reply with either \textless{}ans\text{}X\textless{}/ans\textgreater{} or \textless{}ans\textgreater{}Y\textless{}/ans\textgreater{}.

&You are playing a competition. In each round, you must choose between choices X or Y.

The rules are as follows:
- if both players chooses Y, both players will lose 1 dollar each.
- if both players chooses X, both players will lose 3 dollars each.
- if player A chooses Y and player B chooses X, player A will lose 5 dollars, and player B will lose nothing.
- if player A chooses X and player B chooses Y, player A will lose nothing, and player B will lose 5 dollars.

The competition is played repeatedly against another player. You will be provided with a summary of all previous rounds, showing the actions taken by both you and your opponent and the communication between you two. Based on this information, you need to decide to choose X or Y in the current round.

You are playing as Player A.

The following is a summary of the previous rounds:

No prior rounds.

First, think step by step, using Theory of Mind and planning ahead, what is your action for this round? Then reply with either \textless{}ans\text{}X\textless{}/ans\textgreater{} or \textless{}ans\textgreater{}Y\textless{}/ans\textgreater{}.\\

\end{tabular}
\end{center}
\end{table}

\end{document}